\documentclass[11pt,a4paper]{article}
\usepackage{times,latexsym}
\usepackage{url}
\usepackage[T1]{fontenc}
\usepackage{hyperref}       
\usepackage{booktabs}       
\usepackage{amsfonts}       
\usepackage{nicefrac}       
\usepackage{microtype}      
\usepackage{amsmath}
\usepackage{cleveref}       
\usepackage{lipsum}         
\usepackage{graphicx}
\usepackage{natbib}
\setcitestyle{numbers,square}
\usepackage{doi}
\usepackage{subfig}
\usepackage{algorithmic}
\usepackage{algorithm}
\usepackage{color}
\usepackage{multirow}
\usepackage{arydshln}
\usepackage{enumitem}

\usepackage[acceptedWithA]{tacl2021v1}

\usepackage{xspace,mfirstuc,tabulary}

\newif\iftaclinstructions
\taclinstructionsfalse 
\iftaclinstructions
\renewcommand{\confidential}{}
\renewcommand{\anonsubtext}{(No author info supplied here, for consistency with
TACL-submission anonymization requirements)}
\newcommand{\instr}
\fi

\iftaclpubformat 

\else

\fi

\title{Accurate and Efficient Fine-Tuning of Quantized Large Language Models Through Optimal Balance in Adaptation}

\author{
  Ao Shen, Zhiquan Lai\Thanks{Corresponding author. Trans of the Association for Computational Linguistics 2025, Pre-MIT Press publication.}, Qiang Wang, Xionglve Li, Lizhi Zhang, Dongsheng Li, Jiaxin Li
  \\
  \ \\
  National Key Laboratory of Parallel and Distributed Computing, College of Computer \\Science and Technology, National University of Defense Technology, Changsha, China
  \\
  \texttt{\{shenao,zqlai\}@nudt.edu.cn}
}

\date{}

\begin{document}

\maketitle
\begin{abstract}
  
    Large Language Models (LLMs) have demonstrated impressive performance across various domains. 
    However, the enormous number of model parameters makes fine-tuning challenging, significantly limiting their application and deployment. 
    Existing solutions combine parameter quantization with Low-Rank Adaptation (LoRA), reducing memory usage but causing performance degradation. 
    Additionally, converting fine-tuned models to low-precision representations further degrades performance. 
    In this paper, we identify an imbalance in fine-tuning quantized LLMs with LoRA: overly complex adapter inputs and outputs versus low effective trainability of the adapter, leading to underfitting during fine-tuning.
    Thus, we propose Quantized LLMs fine-tuning with Balanced Low-Rank Adaptation (Q-BLoRA), which simplifies the adapter’s inputs and outputs while increasing the adapter’s rank to alleviate underfitting during fine-tuning. 
    For low-precision deployment, we propose Quantization-Aware fine-tuning with Balanced Low-Rank Adaptation (QA-BLoRA), which aligns with the block-wise quantization and facilitates quantization-aware fine-tuning of low-rank adaptation based on the parameter merging of Q-BLoRA.
    Both Q-BLoRA and QA-BLoRA are easily implemented and offer the following optimizations: (i) Q-BLoRA consistently achieves state-of-the-art accuracy compared to baselines and other variants; (ii) QA-BLoRA enables the direct generation of low-precision inference models, which exhibit significant performance improvements over other low-precision models.
    We validate the effectiveness of Q-BLoRA and QA-BLoRA across various models and scenarios.
    Code will be made available at \href{https://github.com/xiaocaigou/qbaraqahira}{https://github.com/xiaocaigou/qbaraqahira}.
\end{abstract}

\section{Introduction}
Large Language Models (LLMs) \cite{achiam2023gpt,le2022bloom,touvron2023llama,bubeck2023sparks,zhang2023accelerating} have demonstrated remarkable performance across a wide spectrum of Natural Language Processing (NLP) tasks \cite{wei2022emergent}, establishing new benchmarks in various domains.
These models can be fine-tuned for specific applications, significantly enhancing their versatility and adaptability \cite{brown2020language,devlin2018bert,zhao2023survey}. 
However, the fine-tuning process requires substantial memory resources, limiting the accessibility of State-Of-The-Art (SOTA) NLP technologies.

To address this challenge, existing solutions combine parameter quantization with Low-Rank Adaptation (LoRA) fine-tuning \cite{dettmers2024qlora,xu2023qa,qin2024accurate}, leveraging the strengths of both techniques to significantly reduce memory overhead. Initially, Post-Training Quantization (PTQ) \cite{dettmers2023spqr,frantar2023gptq,lin2023awq,xiao2023smoothquant,wei2022outlier,wu2023zeroquant} is applied to convert the pre-trained model into a low-precision representation, thereby reducing the memory required for model loading. Subsequently, LoRA \cite{hu2021lora} is applied for Parameter-Efficient Fine-Tuning (PEFT), introducing a small number of new learnable parameters while keeping most pre-trained parameters fixed, thereby reducing the memory needed for parameter updates.

    \begin{figure*}[htbp]
    	\centering
    	\subfloat[The compraison of Q-BLoRA in the scenario of a 4-bit pre-trained model with a 16-bit adaptation scenario.]{\includegraphics[width=.62\columnwidth,height=.44\columnwidth]{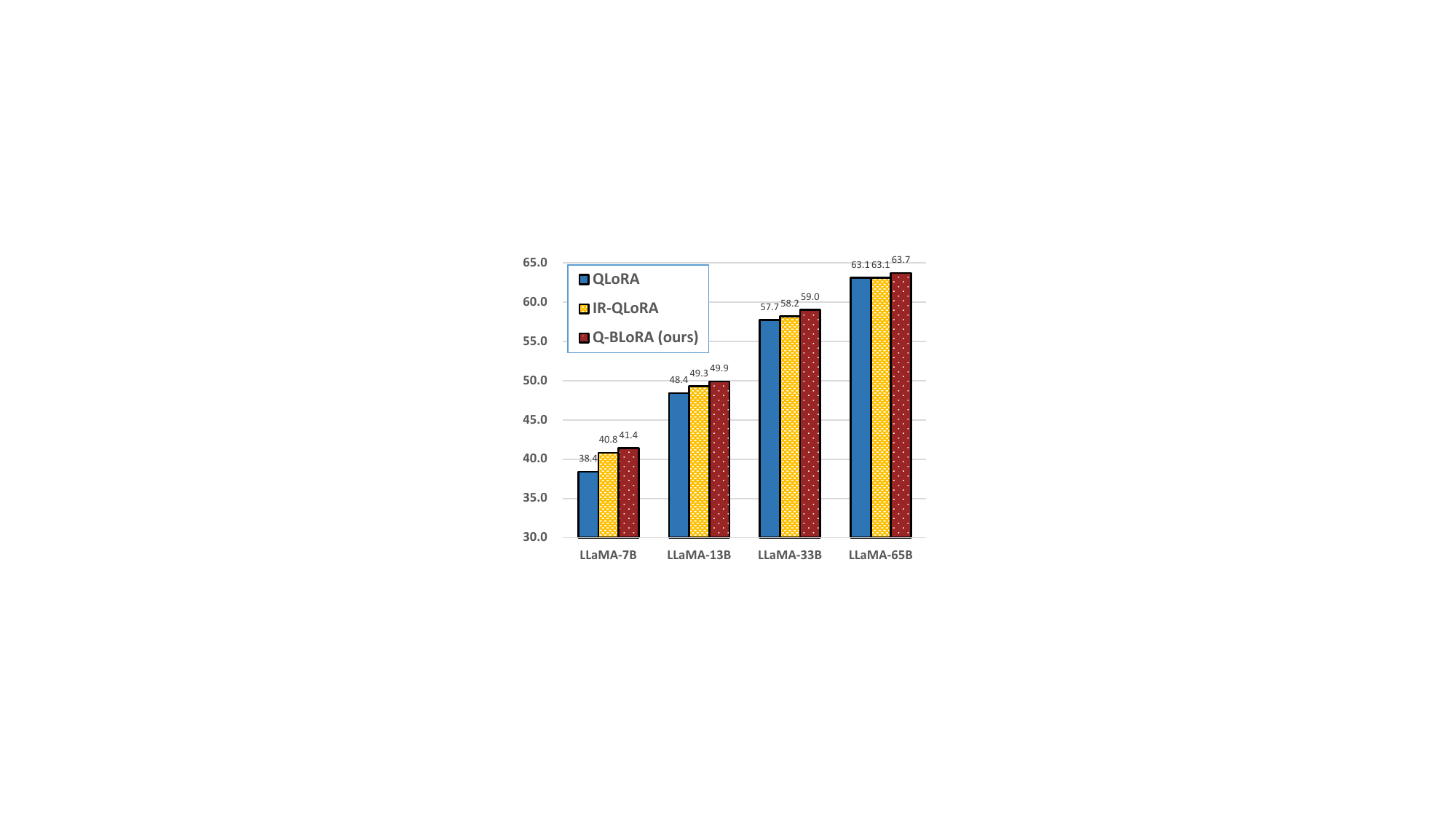}\label{fig_result1}}\hspace{10pt}
    	\subfloat[The compraison of QA-BLoRA in the scenario of a 4-bit inference model scenario.]{\includegraphics[width=.62\columnwidth,height=.44\columnwidth]{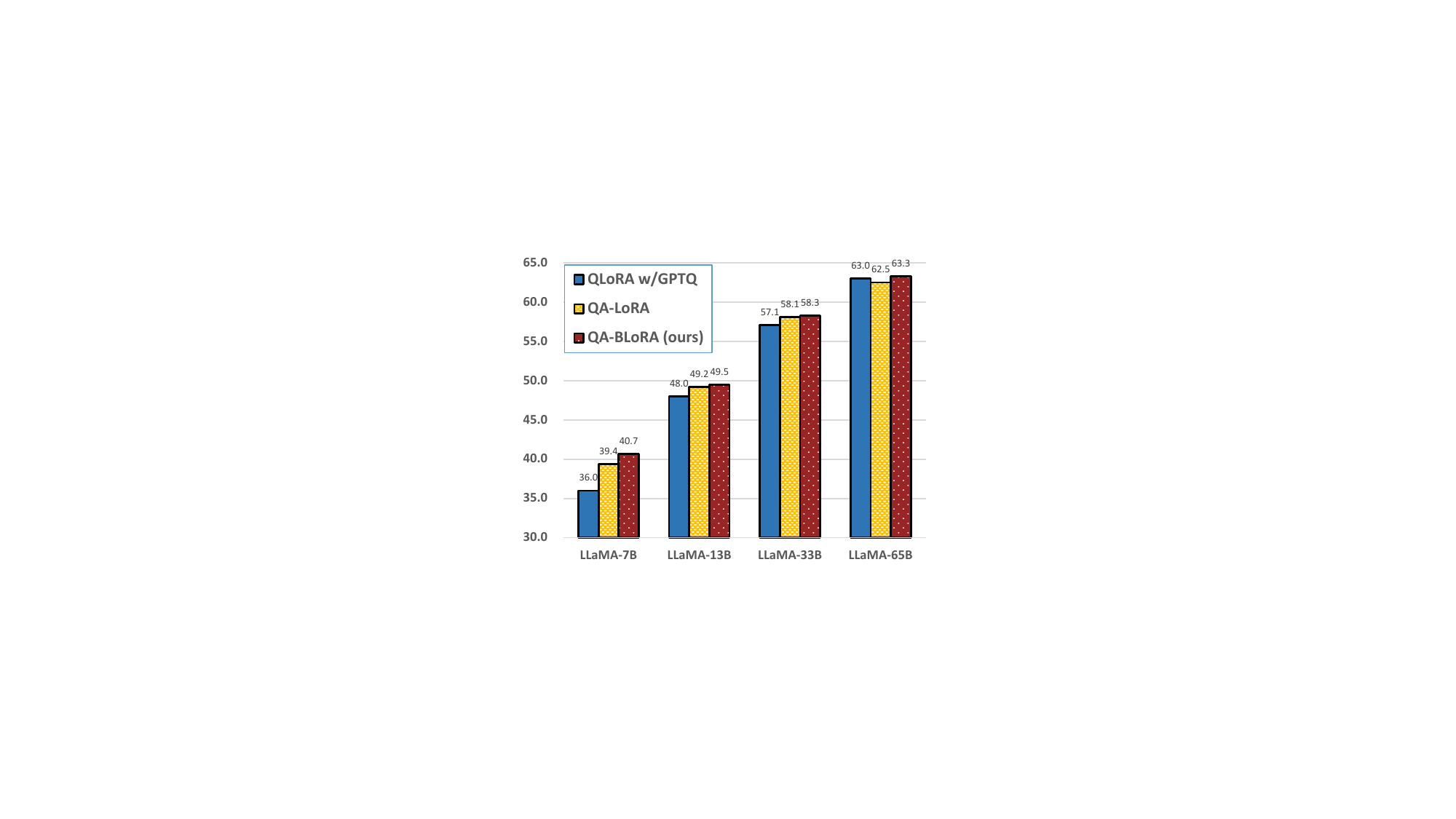}\label{fig_result2}}\hspace{10pt}
        \subfloat[The compraison on LLaMA2 model, where only our Q-BLoRA and QA-BLoRA outperformed their respective baselines.]{\includegraphics[width=.62\columnwidth,height=.44\columnwidth]{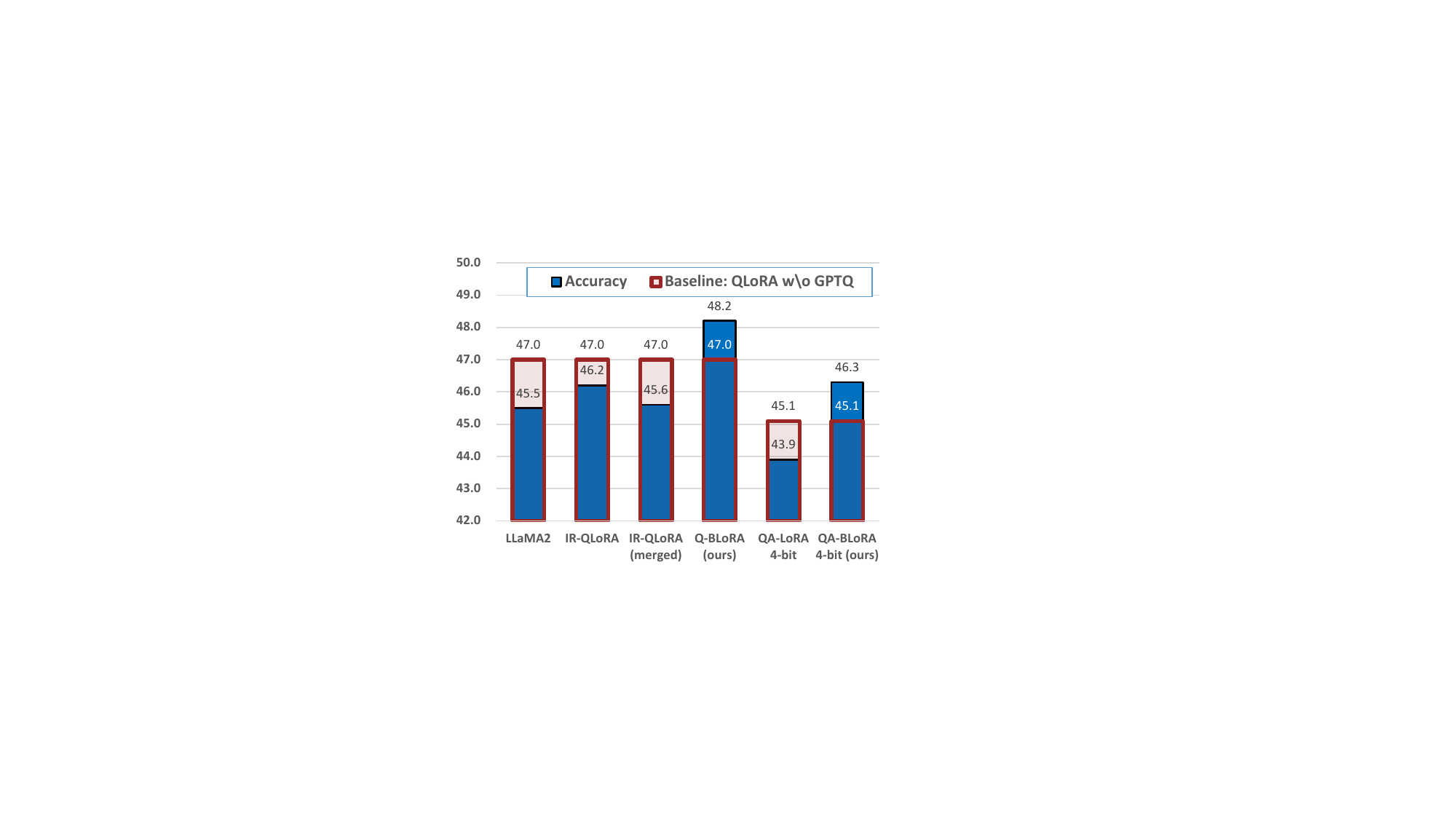}\label{fig_result3}}\\
    	\caption{The comparison of 5-shot MMLU accuracy (\%) based on the LLaMA and LLaMA2 family. All models are fine-tuned on the Alpaca dataset. Full results are provided in Table~\ref{table_main_result_bara} and Table~\ref{table_llama2}.}
            \label{fig_result}
    \end{figure*}

However, despite significantly reducing the memory usage, the combination of parameter quantization and LoRA fine-tuning leads to performance degradation. For example, in the MMLU test, the fine-tuning LLaMA-13B with such approaches leads to an accuracy drop of up to 1.3\% compared to full fine-tuning \cite{dettmers2024qlora}. While recent research attempts to recover lost accuracy through information retention techniques \cite{qin2024accurate}, these approaches often compromise deployment efficiency. Moreover, in low-precision deployment scenarios, compressing the fine-tuned model into a low-precision representation \cite{frantar2023gptq} or directly employing Quantization-Aware Training (QAT) \cite{xu2023qa} requires additional PTQ operations and often struggles to achieve satisfactory performance.

In this paper, we identify an imbalance in fine-tuning quantized LLMs with LoRA: overly complex adapter inputs and outputs versus low effective trainability of the adapter, leading to underfitting during fine-tuning.
To address this issue, we reduce the dimensions of the input and output to simplify the adapter's complexity, while increasing the rank of the low-rank matrices to enhance the adapter's representational capacity.
The compression and restoration of the inputs and outputs are accomplished through non-parameterized operations with negligible computational overhead.
Consequently, we propose Quantized LLMs fine-tuning with Balanced Low-Rank Adaptation (Q-BLoRA), which alleviates the underfitting during fine-tuning and achieves lossless merging of adapter parameters with the pre-trained model.

For low-precision deployment scenarios, we propose Quantization-Aware fine-tuning with Balanced Low-Rank Adaptation (QA-BLoRA), which builds upon the adapter parameter merging of Q-BLoRA. By aligning with the block-wise quantization of the pre-trained model, QA-BLoRA facilitates quantization-aware fine-tuning of low-rank adaptation, enabling the direct generation of low-precision inference models.

Both Q-BLoRA and QA-BLoRA are easy to implement, achieving excellent performance without compromising the original efficiency advantages.
We evaluated our methods on the LLaMA and LLaMA2 model families \cite{touvron2023llama,touvron2023llama2}, as well as the Mistral~\cite{jiang2023mistral} and Gemma~\cite{team2024gemma} models.
Figure~\ref{fig_result1} demonstrates that Q-BLoRA consistently achieves SOTA accuracy compared to other fine-tuning methods using 4-bit pre-trained models with 16-bit adapters.
Figure~\ref{fig_result2} highlights that QA-BLoRA outperforms other 4-bit inference models, even surpassing many benchmarks with 16-bit adapters.
On LLaMA2, as shown in Figure~\ref{fig_result3}, only Q-BLoRA and QA-BLoRA surpass the baseline, showcasing their generalizability across diverse model architectures.
Moreover, Q-BLoRA and QA-BLoRA do not require additional PTQ operations or extra trainable parameters, preserving the efficiency of both the fine-tuning process and the inference model.
Comprehensive evaluations on the MMLU benchmark, question-answering tasks, and evaluations using GPT-4 / ChatGPT confirm the consistent effectiveness of our Q-BLoRA and QA-BLoRA.

\section{Related Work}
\textbf{Quantization of LLMs.} Quantization is a crucial strategy for compressing LLMs, enhancing efficiency and scalability by reducing the bit-width of parameters and activations. Given the high training costs of LLMs, the literature focuses mainly on PTQ, which converts the pre-trained model to lower bit-width representations \cite{dettmers2022gpt3,frantar2023gptq,yao2022zeroquant,wu2023zeroquant,shen2025efficient}. Handling outliers in computation poses a significant challenge, as outliers are critical yet introduce substantial quantization errors. Several approaches address outliers by treating them separately \cite{dettmers2022gpt3, xiao2023smoothquant,wei2022outlier} or by employing calibration data to minimize post-quantization errors \cite{frantar2023gptq, lin2023awq}. However, the performance of quantized models is still affected, and additional computational overhead is often required.

\textbf{LoRA.} PEFT introduces a small set of learnable parameters without updating the majority of the pre-trained parameters, thereby reducing the memory required for updates. Among these, LoRA~\cite{hu2021lora,valipour2022dylora}, as a prevalent PEFT approach, introduces two low-rank matrices and uses their product as an adapter. This method significantly reduces the number of trainable parameters for downstream tasks. Furthermore, after fine-tuning, the adapter parameters can be seamlessly integrated with the pre-trained model, thereby eliminating the need for additional computational overhead of the adapter. Recent studies have further advanced LoRA \cite{hayou2024lora+,jiang2024mora,zi2023delta,zhang2023adaptive,biderman2024lora,tian2024hydralora} and attempted to combine it with model compression techniques \cite{zhang2023loraprune,dettmers2024qlora,mao2024dora}.

\textbf{Combination of LoRA and quantization.} Quantization can compress pre-trained models into low-precision representations, while LoRA reduces the memory required for fine-tuning. Therefore, integrating both approaches allows for the full utilization of their respective advantages.
Typically, QLoRA~\cite{dettmers2024qlora} proposes an efficient fine-tuning method that utilizes 4-bit NormalFloat representations for pre-trained parameters, followed by LoRA-based fine-tuning, significantly reducing memory usage but yielding a suboptimal model. 
QA-LoRA~\cite{xu2023qa} compresses the pre-trained model using GPTQ~\cite{frantar2023gptq} and then fine-tunes it using QAT, allowing the fine-tuned adapter parameters to merge into the low-bit-width representation of the pre-trained model. However, GPTQ requires additional computation, and the QAT method of QA-LoRA introduces significant information loss, leading to degraded model performance. IR-QLoRA~\cite{qin2024accurate} enhances model performance through Information Calibration Quantization (ICQ) for pre-trained model and Information Elastic Connection (IEC) during fine-tuning. Nevertheless, ICQ demands extra computation, and the adapters using IEC are no longer computationally equivalent to the base layer, preventing lossless merging with pre-trained parameters.

\section{The Proposed Approach}
\label{sec_pre}

\subsection{Preliminaries}
\textbf{Baseline: low-bit quantization with low-rank adaptation.}
We begin our notation system with LoRA~\cite{hu2021lora}, which utilizes the product of two matrices to approximate parameter updates during fine-tuning, thus allowing efficient fine-tuning.
Let $\mathbf{W}$ denote the pre-trained parameter matrix of size $D_{in}\times D_{out}$, and $x$ represent the input feature vector of length $D_{in}$.
The output feature $y$ of length $D_{out}$ is computed as $y=\mathbf{W}^{\top} x$.
LoRA introduces two matrices, $\mathbf{A}$ and $\mathbf{B}$, with dimensions $D_{in}\times D_{\mathrm{rank}}$ and $D_{\mathrm{rank}}\times D_{out}$, respectively, where $D_{\mathrm{rank}}\ll min(D_{in},D_{out})$.
This allows the product $\mathbf{AB}$ to form a low-rank matrix of the same size as $\mathbf{W}$ using only a small number of parameters.
During fine-tuning, the computation is adjusted to $y=\mathbf{W}^{\top} x  +  s \cdot \bigtriangleup \mathbf{W}^{\top} x=\mathbf{W}^{\top} x+s \cdot (\mathbf{AB})^{\top} x$, where $s$ is the non-learned coefficient for weight tuning.
The large parameter matrix $\mathbf{W}$ remains frozen, while only $\mathbf{A}$ and $\mathbf{B}$ are updated, significantly reducing the number of trainable parameters.
After fine-tuning, the adapter parameters can be merged with the pre-trained parameters for inference as $\mathbf{W'}=\mathbf{W}+s \cdot \bigtriangleup \mathbf{W}=\mathbf{W}+s\cdot \mathbf{AB}$.

Quantization can build upon LoRA by using low bit-width representations for pre-trained parameters, further reducing the memory required for fine-tuning \cite{dettmers2024qlora}.
A widely adopted quantization method is block-wise quantization \cite{frantar2023gptq,dettmers2024qlora}, which divides the parameter matrix $\mathbf{W}$ into smaller blocks and quantizes each block independently. 
For each block, the quantization can be expressed as:
    \begin{equation}
    \label{eq_quantize}
    \tilde{w}  = \alpha \cdot \widehat{w}  + \beta = \alpha \cdot  \left \lceil \frac{w-\beta }{\alpha }  \right \rfloor    + \beta      ,
    \end{equation}
\noindent where $\widehat{w}$ is the $N$-bit representations of $w$, $\left \lceil \cdot  \right \rfloor$ denotes the rounding operation, which maps the value to the low bit-width representation space.
The parameters $\alpha$ and $\beta$ are a pair of quantization factors along with the block parameters for dequantization.
When applying min-max quantization, $\alpha =  (w_{max} - w_{min}) / (2^N -1)$ and $\beta = w_{min}$.
For absmax quantization, which primarily targets symmetric distributions around zero, $\alpha =  \mathrm{absmax(w)} / (2^N -1)$ and $\beta = 0$.
Combined LoRA and quantization, the calculations of fine-tuning can be expressed as: 
    \begin{small}
        \begin{equation}
        y=\tilde{\mathbf{W}}^{\top} x+s \cdot (\bigtriangleup \mathbf{W})^{\top} x =\tilde{\mathbf{W}}^{\top} x+s \cdot (\mathbf{AB})^{\top} x  ,   \nonumber
        \end{equation}
    \end{small}
\noindent where $\tilde{\mathbf{W}}$ is the quantized parameters of the pre-trained model.
By this approach, fine-tuning of LLMs can be accomplished with a minimal number of GPUs.
However, this fine-tuning approach results in a decrease in accuracy. Additionally, the adapter components $s \cdot \mathbf{AB}$ and pre-trained models $\tilde{\mathbf{W}}$ must utilize high precision (BF16 or FP16) to achieve effective integration. And compressing the fine-tuned model to lower precision further degrades its performance.

\textbf{Our objective: Efficient yet effective adaptation and deployment.}
Our goal is to achieve the following while maintaining the original efficiency:
(i) Obtain high-performance fine-tuned models through LoRA-based fine-tuning of quantized LLMs.
(ii) Seamlessly integrate the adapter parameters into the pre-trained model after fine-tuning, without incurring additional computational overhead during deployment.
Recent works attempt to achieve these two goals, but fail to achieve both, even with additional computation.
Specifically, QLoRA~\cite{dettmers2024qlora} inherits the deployment advantages of LoRA by enabling lossless adapter integration. 
However, it suffers from significant performance degradation.
IR-QLoRA~\cite{qin2024accurate} aims to achieve better performance through fine-tuning.
However, the enhancement of accuracy relies on additional computations and trainable parameters. 
Moreover, the shortcut-like structure modifies the computational graph, preventing the adapter from effectively integrating with the pre-trained model.
QA-LoRA~\cite{xu2023qa} proposes a QAT fine-tuning method that allows the side weights $s \cdot \mathbf{AB}$ to merge into the quantized pre-trained model $\tilde{\mathbf{W}}$.
However, the quantization of pre-trained model \cite{frantar2023gptq} require the computation for calibration, and the accuracy of the fine-tuned model remains far from the limits regarding accuracy.

Therefore, we propose fine-tuning methods for quantized LLMs that are tailored to both high-precision floating-point deployment scenarios (e.g., GPUs) and low-precision deployment scenarios (e.g., edge devices). Our approach aims to simultaneously achieve high model performance and efficient deployment, while maintaining the high memory efficiency of the fine-tuning process.


\subsection{Q-BLoRA}
\label{sec_bara}

    \begin{figure*}[htbp]
    	\centering
    	\subfloat[Illustration of gradient and accuracy trends under ideal conditions.]{\includegraphics[width=.61\columnwidth,height=.52\columnwidth]{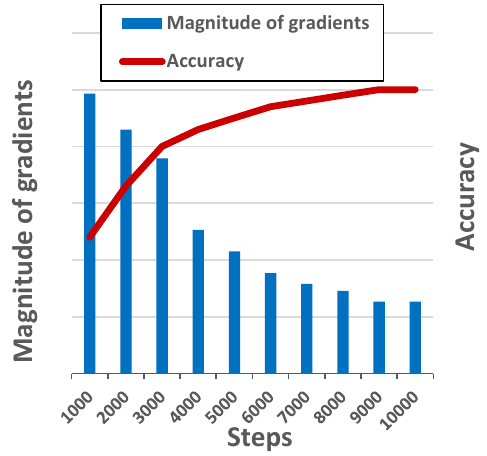}\label{fig_underfiit1}}\hspace{3pt}
    	\subfloat[Gradient and accuracy changes during LoRA fine-tuning of quantized LLaMA2-7B.]{\includegraphics[width=.74\columnwidth,height=.52\columnwidth]{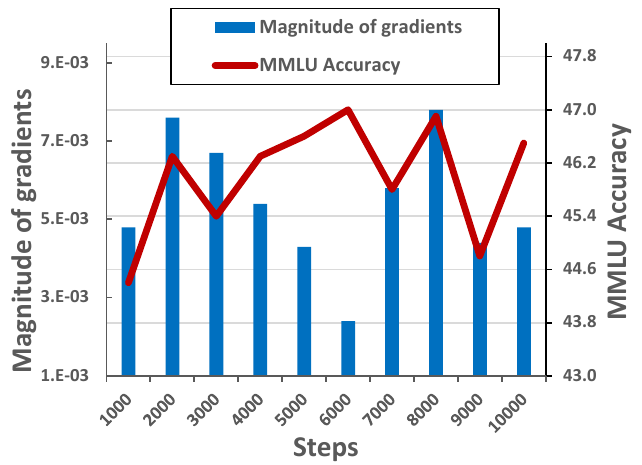}\label{fig_underfiit2}}\hspace{3pt}
        \subfloat[Loss changes of different methods during fine-tuning of LLaMA2-7B.]{\includegraphics[width=.61\columnwidth,height=.52\columnwidth]{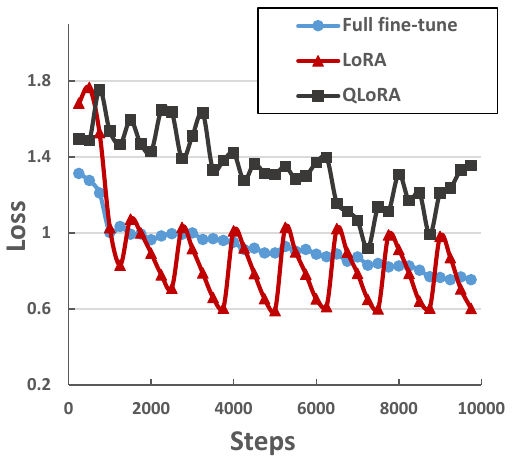}\label{fig_underfiit3}}\\
        \subfloat[Magnitude of the input, output and weights of a attn.q\_proj layer in Llama2-7B.]{\includegraphics[width=.78\columnwidth,height=.54\columnwidth]{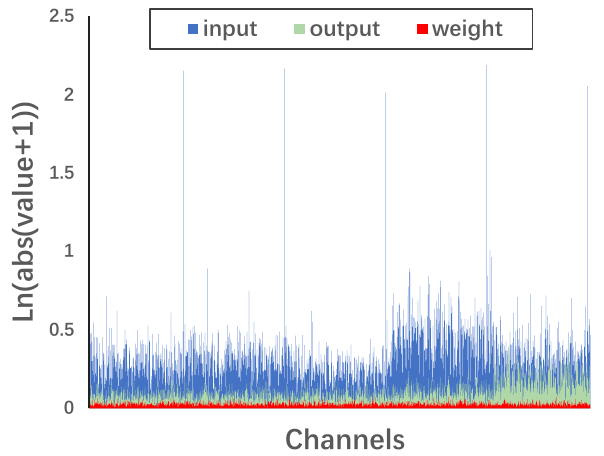}\label{fig_underfiit4}}\hspace{10pt}
        \subfloat[Gradient and accuracy changes during LLaMA2 fine-tuning with different balancing factors.]{\includegraphics[width=1.05\columnwidth,height=.54\columnwidth]{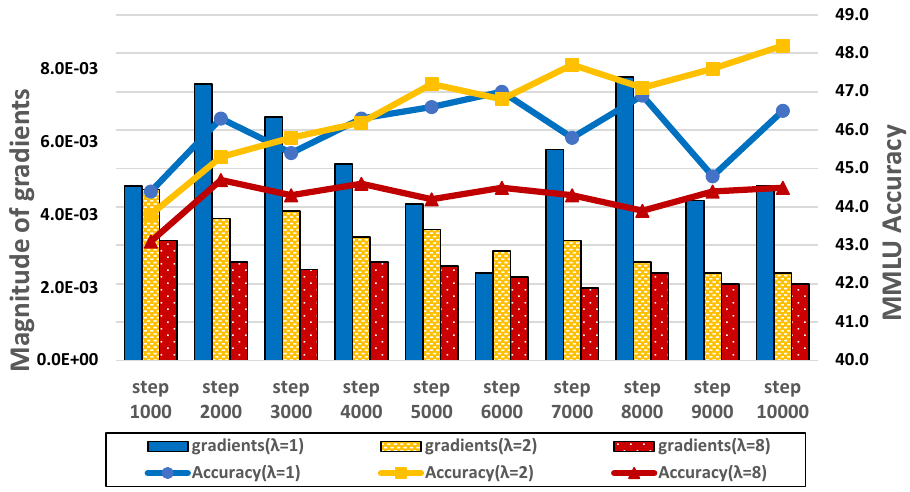}\label{fig_underfiit5}}\\
    	\caption{Analysis of fine-tuning with low-rank adaptation in quantization scenario.}
            \label{fig_underfit}
    \end{figure*}

\textbf{The imbalance in fine-tuning quantized LLMs with LoRA.} 
We identify an imbalance in fine-tuning quantized LLMs with LoRA: overly complex adapter inputs and outputs versus low effective trainability of the adapter, leading to underfitting during fine-tuning.

Typically, neural networks require sufficient or even excessive representational capacity to progressively fit and learn knowledge through continuous training. When the model’s representational capacity is adequate, the training loss steadily decreases, parameters gradually converge, gradients diminish, and accuracy improves, as shown in Figure~\ref{fig_underfiit1}.
However, when quantized LLMs are fine-tuned using LoRA, as shown in Figure~\ref{fig_underfiit2}, significant gradients persist even in the later stages of training, and accuracy exhibits noticeable fluctuations. These observations suggest that fine-tuning quantized LLMs with LoRA may result in underfitting\footnote{In the case of the Q, K, and V layers in LLaMA2-7B, the rank of the adapter matrices is typically only 16 to 64 \cite{dettmers2024qlora}, while the input and output vectors have a dimensionality as high as 4096. This contrasts sharply with conventional training/fine-tuning methods, where, during full fine-tuning of LLaMA2-7B, the number of trainable parameters in the Q, K, and V layers is 4096 times that of their input and output dimensions.}.

As shown in Figure~\ref{fig_underfiit3}, compared to full fine-tuning, the loss when fine-tuning quantized LLMs with LoRA (i.e., QLoRA) remains consistently high and fluctuates, indicating that the model struggles to fit. In contrast, when fine-tuning a non-quantized LLM using LoRA (i.e., LoRA), the loss, despite some fluctuations, decreases to a much lower level. This discrepancy arises because, in LoRA, the pre-trained model has already acquired sufficient knowledge, requiring only minimal adaptation during fine-tuning to fit task-specific knowledge. However, in QLoRA, the quantization process introduces information loss, forcing the adapter to not only learn task-specific knowledge but also compensate for the information lost during quantization. This dual burden significantly increases the difficulty of the fine-tuning process, resulting in underfitting.

The above findings indicate that fine-tuning quantized LLMs with LoRA leads to underfitting, which adversely affects the model's performance.

\textbf{The Rise of Q-BLoRA.} 
Recent studies have proposed potential solutions to address this imbalance problem. Specifically, there is a certain degree of redundancy in the input and output of the layers in LLM models \cite{xiao2023smoothquant, dettmers2022gpt3, wu2023zeroquant}, and not all features from every channel are equally important for the model's performance. Additionally, due to the significant numerical differences across channels (as shown in Figure~\ref{fig_underfiit4}), in cases of underfitting, excessive parameter updates may even hinder model convergence, leading to performance degradation.

We use $\lambda$ as the balancing factor, which serves as the compression factor for both the input and output dimensions and as the multiplier for rank expansion.
We record the accuracy and gradient values during fine-tuning for various $\lambda$.
As shown in Figure~\ref{fig_underfiit5}, when $\lambda=1$, the input, output and rank of the adapter remain unmodified. In this case, the gradients during fine-tuning remain consistently large, and the accuracy exhibits significant fluctuations. When $\lambda=2$, the gradients are noticeably reduced and show a decreasing trend, while the accuracy improves. This suggests that the fine-tuning process achieves a better balance, effectively mitigating the underfitting issue. Finally, when $\lambda=8$, the gradients decrease further, indicating a substantial reduction in the difficulty of fitting. However, at this stage, excessive compression of the input and output dimensions leads to a loss of valuable information, resulting in performance degradation.

These observations indicate that simplifying the adapter's input and output dimensions while expanding its rank can optimize the balance during fine-tuning, alleviating the underfitting problem. 

    \begin{figure*}[t]
    \begin{center}
    \includegraphics[width=0.99\textwidth,height=0.4\textwidth]{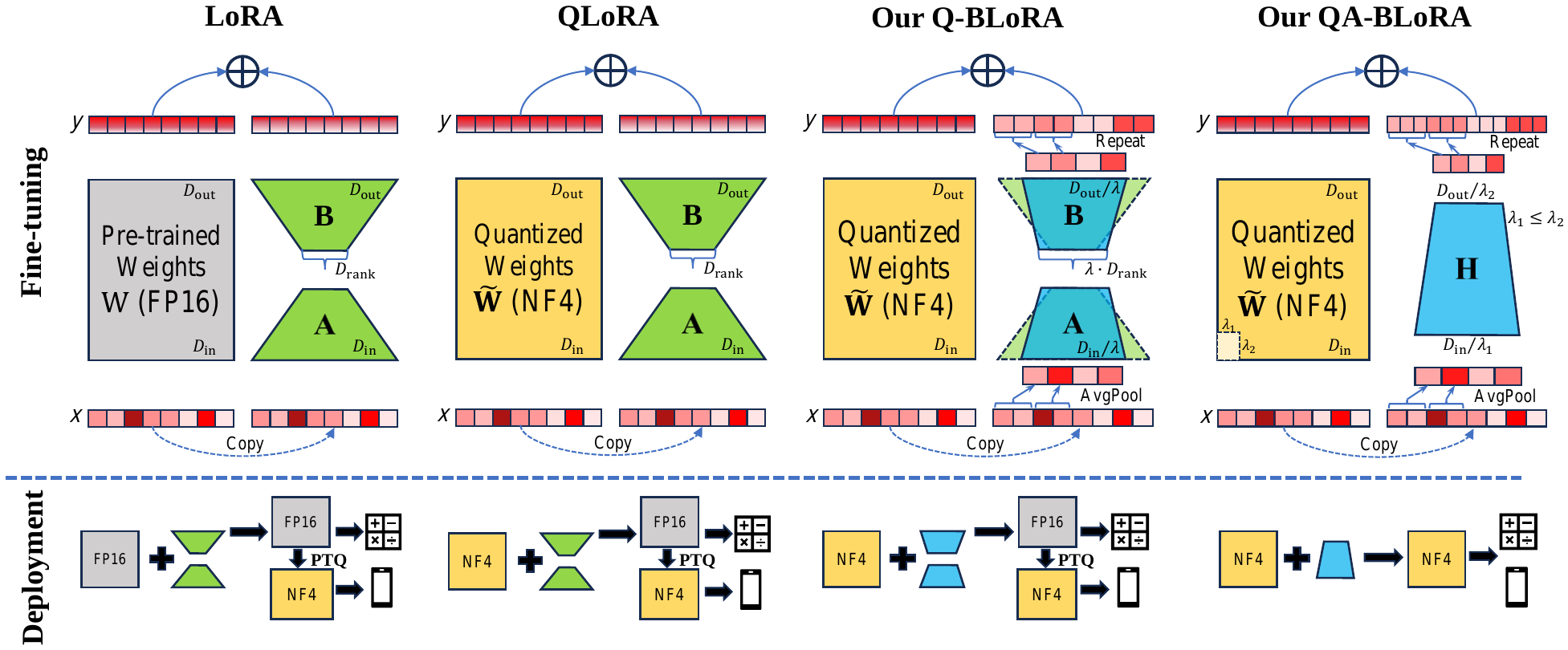}
    \end{center}
    \caption{An illustration of our Q-BLoRA and QA-BLoRA. Compared to LoRA and QLoRA, our Q-BLoRA achieves a better fine-tuning balance by increasing the rank of adaptation and using non-parameter operators to change the dimension of input and output. Our QA-BLoRA employs a single matrix for adaptation, enabling the fine-tuning of an efficient model ready for inference.}
    \label{fig_bara_hira}
    \end{figure*}

Based on this analysis, we propose Quantized LLMs fine-tuning with Balanced Low-Rank Adaptation (Q-BLoRA), as illustrated in Figure~\ref{fig_bara_hira}. 
Considering the parameter distribution of LLMs, Q-BLoRA employs NF4 to compress the pre-trained model \cite{dettmers2023case}. On the adapter side, Q-BLoRA first applies the \textsf{AvgPool($\lambda$)} operation to the input $x$, transforming it into $x'$ with dimensions $D_{in}/\lambda$. The dimensions of the matrices $\mathbf{A}$ and $\mathbf{B}$ in LoRA are subsequently adjusted to $(\frac{D_{in}}{\lambda}, \lambda \cdot D_{\mathrm{rank}})$ and $(\lambda \cdot D_{\mathrm{rank}}, \frac{D_{out}}{\lambda})$, respectively. The output $y'$ with dimension $D_{out}/\lambda$ is computed as $y' = (\mathbf{AB})^{\top} x'$.
Next, the output $y'$ undergoes the \textsf{repeat\_interleave($\lambda$)} operation, repeating elements and producing the final adapter output $y$, with dimensions restored to $D_{out}$.

In this process, Q-BLoRA reduces the input and output dimensions of the adapter by a factor of $\lambda$, which not only reduces the complexity of fine-tuning but also alleviates fluctuations in the adapter's input and output, mitigating the negative effects of excessive parameter updates. Simultaneously, it expands the rank of the adapter to enhance its effective representational capacity. By achieving an optimal balance between input/output dimensions and rank, Q-BLoRA mitigates underfitting without increasing the number of trainable parameters. For a detailed analysis of the choice of $\lambda$, refer to Section~\ref{sec_ablation}.

\textbf{Merging Q-BLoRA into pre-trained model.} 
A notable advantage of LoRA~\cite{hu2021lora} and QLoRA~\cite{dettmers2024qlora} is that the low-rank adaptation $\bigtriangleup \mathbf{W}$ has the same dimensions as the pre-trained model $\mathbf{W}$ or $\tilde{\mathbf{W}}$. This allows for merging into the pre-trained model, making inference no longer require additional computation from the adapter. However, the adaptation weight of our Q-BLoRA $\bigtriangleup \mathbf{W}_{BLoRA}$ is in shape $(D_{in}/\lambda, D_{out}/\lambda)$, necessitating the design of methods to integrate Q-BLoRA effectively.

Let's start by analyzing the calculation when Q-BLoRA is \textit{just compressing the input}, and denote the balancing factor at this stage as $\lambda _1$. 
The \textsf{AvgPool($\lambda _1$)} operation changes $x=(x_1,x_2,...,x_{D_{in}})^\top $ of length $D_{in}$ to $x^{pool}=(\frac{ {\textstyle \sum_{i=1}^{\lambda _1}}x_i }{\lambda  _1} ,    \frac{ {\textstyle \sum_{i=\lambda _1+1}^{2\lambda _1}}x_i }{\lambda  _1} ,...,     \frac{ {\textstyle \sum_{i=D_{in}-\lambda  _1+1}^{D_{in}}}x_i }{\lambda  _1} )^\top$ of length $D_{in}/\lambda _1$, and the adaptation weight $\bigtriangleup \mathbf{W}_{BLoRA}^{in}$ is in shape $(D_{in}/\lambda _1, D_{out})$.
The computation can be described as Equation~\eqref{eq_chengfa}.
\begin{figure*}
    \begin{small}
\begin{align}
    (\bigtriangleup \mathbf{W}_{BLoRA}^{in})^\top  & x^{pool} =            
    \begin{pmatrix}
     w^{in}_{11} & w^{in}_{21} & ... & w^{in}_{\frac{D_{in}}{\lambda_1 }, 1} \\
     w^{in}_{12} & w^{in}_{22} & ... & w^{in}_{\frac{D_{in}}{\lambda_1 }, 2}\\
     \vdots  & \vdots & \ddots  & \vdots\\
     w^{in}_{1,D_{out}} & w^{in}_{2,D_{out}} & ... & w^{in}_{\frac{D_{in}}{\lambda_1 },D_{out}}
    \end{pmatrix}
    \begin{pmatrix}
     \frac{ {\textstyle \sum_{i = 1}^{\lambda_1}}x_i }{\lambda_1 }\\
    \frac{ {\textstyle \sum_{i = \lambda_1+1}^{2\lambda_1}}x_i }{\lambda_1 } \\
    \vdots \\
    \frac{ {\textstyle \sum_{i = D_{in}-\lambda_1 +1}^{D_{in}}}x_i }{\lambda_1 } 
    \end{pmatrix}     \label{eq_chengfa}\\
    =
 &   \frac{1}{\lambda_1 }   \begin{pmatrix}
     {w^{in}_{11}\textstyle \sum_{i = 1}^{\lambda_1}}x_i +      w^{in}_{21}{\textstyle \sum_{i = \lambda_1+1}^{2\lambda_1}}x_i              +  \cdots +  w^{in}_{\frac{D_{in}}{\lambda_1 } ,1}{\textstyle \sum_{i = D_{in}-\lambda_1 +1}^{D_{in}}}x_i \\
     \vdots \\
    {w^{in}_{1,D_{out}}\textstyle \sum_{i = 1}^{\lambda_1}}x_i +           w^{in}_{2,D_{out}}{\textstyle \sum_{i = \lambda_1+1}^{2\lambda_1}}x_i   +      \cdots +  w^{in}_{\frac{D_{in}}{\lambda_1 }, D_{out}}{\textstyle \sum_{i = D_{in}-\lambda_1 +1}^{D_{in}}}x_i 
    \end{pmatrix}          \nonumber 
    \end{align}
    \end{small}
\end{figure*}
Equation~\eqref{eq_chengfa} can be equivalent to the baseline where the multiplication is between $\bigtriangleup \mathbf{W}$ in shape $(D_{in}, D_{out})$ and $x$ of length $D_{in}$, under the condition that $w_{m,n}^{in} = \lambda_1 w_{\lambda_1 m+1, n} = \lambda_1 w_{\lambda_1 m+2, n} = \cdots = \lambda_1 w_{\lambda_1 m+\lambda_1, n}$ for any $m\in [0,D_{in}/\lambda_1) \cap \mathbb{Z} $ and $n\in [0,D_{out}] \cap \mathbb{Z} $ is satisfied.

When \textit{only the Q-BLoRA output is considered}, we get the output $y^{re} = (y^{re}_1, y^{re}_2, \cdots, y^{re}_{D_{out}/\lambda_2})$ of length $D_{out}/\lambda_2$. Then, the \textsf{repeat\_interleave} operation expands $y^{re}$ to $y = (y^{re}_1,y^{re}_1,\cdots,y^{re}_1,y^{re}_2,y^{re}_2,\cdots,y^{re}_2,\cdots,y^{re}_{D_{out}/\lambda_2})$ of length $D_{out}$.
Similarly, it can be equivalent to the baseline when $w_{m,n}^{out} = w_{m,\lambda_2 n+1} = w_{m,\lambda_2 n+2} = \cdots = w_{m,\lambda_2 n+\lambda_2}$ for any $m\in [0,D_{in}) \cap \mathbb{Z} $ and $n\in [0,D_{out}/\lambda_2) \cap \mathbb{Z} $.

Together, we can get the $\bigtriangleup \mathbf{W}$ in shape $(D_{in}, D_{out})$ which is equivalent to the fine-tuning using Q-BLoRA. This equivalent $\bigtriangleup \mathbf{W}$ can be divided into $\frac{D_{in} D_{out}}{\lambda_1 \lambda_2}$ blocks of size $(\lambda_1 , \lambda_2)$, each with the same value of elements. 
Our Q-BLoRA can be implemented by inserting/modifying a few lines of code, as described in Algorithm~\ref{al_BLoRA}.

    \begin{algorithm*}[!h]
        \caption{Q-BLoRA Pseudocode in the PyTorch-like style}
        \label{al_BLoRA}

        \begin{algorithmic}
            
            \begin{small}

            \STATE \hspace{-4mm} \textcolor{blue}{\# lambda: the balancing factor; W\_hat: the quantized weight; scaling: the scaling factor for adaptation}\
            \STATE \hspace{-4mm} \textcolor{blue}{\# D\_in, D\_out: the input and output dimensions of the pretrained model;  D\_rank: the dimension of low-rank adaptation of QLoRA}\
            \STATE \hspace{-4mm} \textcolor{red} {qblora\_pool = nn.AvgPool1d(lambda)}
            \STATE \hspace{-4mm} \textcolor{red}{qblora\_A = nn.Linear(lambda * D\_rank, D\_in / lambda)}
            \STATE \hspace{-4mm} \textcolor{red}{qblora\_B = nn.Linear(D\_out / lambda, lambda * D\_rank)}
            \STATE \hspace{-4mm} def qblora\_forward(x, W\_hat, alpha, beta, qblora\_A, qblora\_B):
                \STATE \hspace{1mm} W\_tilde = dequantize(W\_hat, alpha, beta)
                \STATE \hspace{1mm} result = x \texttt{@}  W\_tilde
                \STATE \hspace{1mm} result += \textcolor{red}{qblora\_pool(x)} \texttt{@} qblora\_A.transpose(0, 1) \texttt{@} qblora\_B.transpose(0, 1)
                \STATE \hspace{1mm} result = scaling * \textcolor{red}{torch.repeat\_interleave(result, lambda, dim = 2)}
                \STATE \hspace{1mm} return result
            \STATE \hspace{-4mm} def get\_delta\_weight(qblora\_A, qblora\_B):
                \STATE \hspace{1mm} delta\_weight = torch.repeat\_interleave(qblora\_A \texttt{@} qblora\_B, lambda, dim=0), lambda, dim=1) / lambda
                \STATE \hspace{1mm} delta\_weight = torch.repeat\_interleave(delta\_weight, lambda, dim=1)
                \STATE \hspace{1mm} return delta\_weight

            

            \end{small}
        \end{algorithmic}
    \end{algorithm*}

\subsection{QA-BLoRA}

    \begin{figure*}[t]
    \begin{center}
    \includegraphics[width=0.99\textwidth,height=0.54\textwidth]{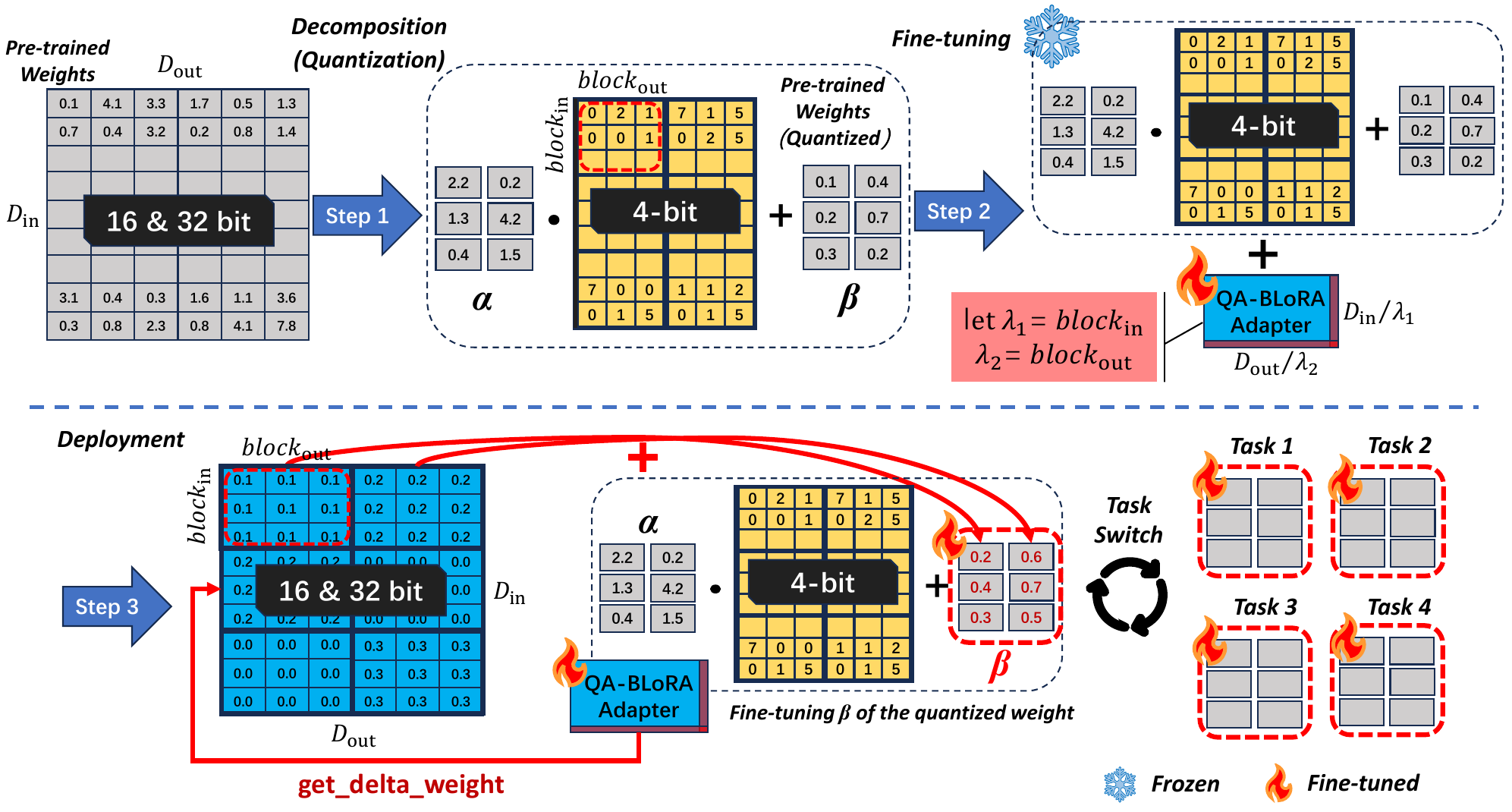}
    \end{center}
    \caption{Quantization-aware fine-tuning of LoRA for low-precision deployment and task switching.}
    \label{fig_finetune_hira}
    \end{figure*}

In some application scenarios, low-precision models are required for inference without relying on high-precision formats such as FP16 or BF16. To address this, Hugging Face\footnote{\url{https://huggingface.co/blog/overview-quantization-transformers}} recommends PTQ after fine-tuning. While this approach is feasible, PTQ inevitably leads to some accuracy degradation. In contrast, QAT adapts to low precision during the training process, thereby improving performance. Therefore, the optimal solution is to directly obtain a low-precision inference model through quantization-aware fine-tuning.

Since the quantized pre-trained model is represented using a triplet $(\hat{w}, \alpha, \beta)$ in a block-wise fashion, as shown in Equation~\eqref{eq_quantize}, where $\hat{w}$ is a discrete series corresponding to the rounding operation $\left \lceil \cdot \right \rfloor$, and $\alpha$ and $\beta$ are high-precision floating-point numbers, considered continuous. For the adapter weight $\bigtriangleup \mathbf{W}$ to merge into the quantized pre-trained model, the values of $(\bigtriangleup w - \beta) / \alpha$ within each block must align with the discrete series defined by the rounding operation. Meeting this condition is challenging, especially since each block corresponds to different parameters $\alpha$ and $\beta$.

Thus, we need to relax the constraints. A feasible solution is to make all $\bigtriangleup w$ within each block equal, allowing $\bigtriangleup w$ to be viewed as an adjustment to $\beta$, thereby enabling quantization-aware fine-tuning of LoRA. As analyzed in Section~\ref{sec_bara}, this can be easily achieved through the parameter merging of Q-BLoRA, wherein every block of size $(\lambda_1 , \lambda_2)$ within the $\bigtriangleup \mathbf{W}$ of Q-BLoRA has equal elements. We only need to adjust $\lambda _1$ and $\lambda _2$ so that it aligns to the size of the quantization block. The detailed process is shown in Figure~\ref{fig_finetune_hira}.

    \begin{algorithm*}[!h]
        \caption{QA-BLoRA Pseudocode in the PyTorch-like style}
        \label{al_hira}
        
        \begin{algorithmic}
            \begin{small}

            \STATE \hspace{-4mm} \textcolor{blue}{\# lambda\_1, lambda\_2: the compressing factor of input \& output; W\_hat: the quantized weight}\
            \STATE \hspace{-4mm} \textcolor{blue}{\# D\_in, D\_out: the input \& output dimensions of the pretrained model; scaling: the scaling factor for adaptation}\
            \STATE \hspace{-4mm} \textcolor{red} {qablora\_pool = nn.AvgPool1d(lambda\_1)}
            \STATE \hspace{-4mm} \textcolor{red}{qablora\_H = nn.Linear(D\_out / lambda\_2, D\_in / lambda\_1)}
            \STATE \hspace{-4mm} def qablora\_forward(x, W\_hat, alpha, beta, qablora\_H):
                \STATE \hspace{1mm} W\_tilde = dequantize(W\_hat, alpha, beta)
                \STATE \hspace{1mm} result = x \texttt{@}  W\_tilde
                \STATE \hspace{1mm} result += \textcolor{red}{qablora\_pool(x)} \texttt{@} qablora\_H.transpose(0, 1)
                \STATE \hspace{1mm} result = scaling * \textcolor{red}{torch.repeat\_interleave(result, lambda\_2, dim = 2)}
                \STATE \hspace{1mm} return result
            \STATE \hspace{-4mm} def get\_new\_beta(qablora\_H, beta):
                \STATE \hspace{1mm} new\_beta = torch.repeat\_interleave(qablora\_A, lambda\_1, dim=0) / lambda\_1
                \STATE \hspace{1mm} new\_beta = scaling * torch.repeat\_interleave(delta\_weight, lambda\_2, dim=1) + beta
                \STATE \hspace{1mm} return new\_beta

            \end{small}
        \end{algorithmic}
    \end{algorithm*}

Based on this analysis, we propose Quantization-Aware fine-tuning with Balanced Low-Rank Adaptation (QA-BLoRA). This approach compresses the adapter's inputs and outputs to align with block-wise quantization,
thereby achieving quantization-aware fine-tuning for LoRA. 
Typically, pre-trained models use block quantization with sizes of $4 \times 8$ or $8 \times 8$ \cite{frantar2023gptq,dettmers2023case}. When both the input and output are significantly simplified, we employ a single matrix as the adapter to enhance the fine-tuning capability \cite{jiang2024mora}. 
Furthermore, as shown in Figure~\ref{fig_underfiit4}, the input contains more information (including more outliers) than the output. Consequently, we set $\lambda_1 \leq \lambda_2$, prioritizing the preservation of information in the input.
For an explanation of QA-BLoRA, see Figure~\ref{fig_bara_hira}. For further analysis of $\lambda_1$ and $\lambda_2$, see Section~\ref{sec_ablation}.
Our QA-BLoRA can also be implemented by inserting/modifying a few lines of code, as described in Algorithm~\ref{al_hira}.

\section{Experiments}

\subsection{Settings}
We establish Q-BLoRA and QA-BLoRA on the LLaMA and LLaMA2 model families \cite{touvron2023llama,touvron2023llama2}, the Mistral~\cite{jiang2023mistral} and Gemma~\cite{team2024gemma} models, using Alpaca~\cite{taori2023stanford} and Flan v2~\cite{longpre2023flan} datasets.
For evaluation, we utilize the Massively Multitask Language Understanding (MMLU)~\cite{hendrycks2020measuring}, the zero-shot CommonsenseQA benchmarks (e.g.HellaSwag~\cite{zellers2019hellaswag}, PIQA~\cite{bisk2020piqa}, WinoGrande~\cite{sakaguchi2021winogrande}, ARC~\cite{clark2018think}, BoolQ~\cite{clark2019boolq}, and OpenBookQA~\cite{mihaylov2018can}) and evaluations using GPT-4 / ChatGPT (including Vicuna~\cite{chiang2023vicuna}, Koala~\cite{vu2023koala}, WizardLM~\cite{xu2023wizardlm}, Self-instruct~\cite{wang2023self} and LIMA~\cite{zhou2024lima}). We follow the official code and configuration of QLoRA~\cite{dettmers2024qlora}, IR-QLoRA~\cite{qin2024accurate}, QA-LoRA~\cite{xu2023qa}, PEQA~\cite{kim2024memory}, and GPTQ~\cite{achiam2023gpt}. 

\subsection{Main Results and Efficiency}
\textbf{Comparison against recent competitors of fine-tuning LLaMA for MMLU.}
Table~\ref{table_main_result_bara} compares our Q-BLoRA and QA-BLoRA with SOTA LoRA-based quantization methods, including QLoRA~\cite{dettmers2024qlora}, QA-LoRA~\cite{xu2023qa}, and IR-QLoRA~\cite{qin2024accurate}. We also compare with PEQA~\cite{kim2024memory}, which does not use LoRA but fine-tunes the quantized scaling instead. 
The "4+16" bit width indicates the inference model requires a 4-bit pre-trained model togethor with a 16-bit adapter, while "4+16\&16" signifies the pre-trained model and adapter can be merged into a single 16-bit model, thereby enhancing inference efficiency.

Comprehensive results in Table~\ref{table_main_result_bara} demonstrate that Q-BLoRA and QA-BLoRA consistently outperform all comparative methods by a convincing margin in their respective application scenarios.
Q-BLoRA achieves the SOTA accuracy across models of various sizes. Compared to the most related method, QLoRA, our Q-BLoRA shows a significant improvement in accuracy. When compared to IR-QLoRA, the most competitive method in terms of accuracy, Q-BLoRA still achieves higher accuracy without requiring extra optimization for pre-trained model quantization or introducing extra trainable parameters. Moreover, IR-QLoRA alters the adapter's computational graph, preventing lossless merging with the pre-trained model and degrading performance after merging, whereas Q-BLoRA supports lossless merging.

QA-BLoRA can directly produce a 4-bit inference model through fine-tuning, making it more suitable for deployment on edge devices. The results show that QA-BLoRA has a significant accuracy advantage over other low-precision methods, whether QLoRA fine-tuning followed by GPTQ compression or the recent QAT method QA-LoRA. It even surpasses most fine-tuned 16-bit inference models in accuracy. Notably, our QA-BLoRA also does not require extra optimization for pre-trained model quantization or extra trainable parameters.

    \begin{table*}
    \centering
    \caption{Accuracy (\%) comparison of LLaMA on the MMLU fine-tuned on Alpaca dataset. }
    \label{table_main_result_bara}
    \resizebox{.94\textwidth}{!}{
    \setlength{\tabcolsep}{5mm}{
    \begin{tabular}{ccccccccc} 
    \toprule
    \multirow{2}{*}{Method~} & \multirow{2}{*}{\begin{tabular}[c]{@{}c@{}}\#Bits~\\\end{tabular}} & \multirow{2}{*}{\begin{tabular}[c]{@{}c@{}}No extra~\\PTQ\end{tabular}} & \multirow{2}{*}{\begin{tabular}[c]{@{}c@{}}No extra~\\\#params\end{tabular}} & \multicolumn{5}{c}{MMLU}                                       \\ 
    \cline{5-9}
                             &                                                                    &                                                                           &                                                                                & ~ Hums.~~ & ~ STEM~~ & ~ Social~~ & ~ Other~~ & ~ ~Avg.~ ~     \\ 
    \midrule
    LLaMA-7B                 & 16                                                                 & -                                                                         & -                                                                              & 33.3      & 29.8     & 37.8       & 38.0      & 34.6           \\ 
    \midrule
    QLoRA~~                  & ~4+16\&16~                                                         & Yes                                                                       & Yes                                                                            & 36.1      & 31.9     & 42.0       & 44.5      & 38.4           \\
    IR-QLoRA~~               & 4+16                                                               & No                                                                        & No                                                                             & 38.6      & 34.6     & 45.2       & 45.5      & 40.8           \\
    IR-QLoRA(merged)        & 16                                                                 & No                                                                        & No                                                                             & 37.6      & 33.2     & 44.3       & 44.7      & 39.4           \\
    \textbf{Q-BLoRA (ours)}   & 4+16\&16                                                           & \textbf{Yes}                                                              & \textbf{Yes}                                                                   & 43.4      & 36.1     & 42.1       & 46.0      & \textbf{41.4}  \\ 
    \midrule
    PEQA                     & 4                                                                  & Yes                                                                       & Yes                                                                            & 34.9      & 28.9     & 37.5       & 40.1      & 34.8           \\
    QLoRA w/ GPTQ            & 4                                                                  & No                                                                        & Yes                                                                            & 33.8      & 31.3     & 37.4       & 42.2      & 36.0           \\
    QA-LoRA                  & 4                                                                  & No                                                                        & Yes                                                                            & 36.6      & 32.4     & 44.8       & 44.9      & 39.4           \\
    \textbf{QA-BLoRA (ours)}   & 4                                                                  & \textbf{Yes}                                                              & \textbf{Yes}                                                                   & 43.3      & 36.0     & 40.8       & 44.3      & \textbf{40.7}  \\ 
    \midrule
    LLaMA-13B                & 16                                                                 & -                                                                         & -                                                                              & 44.0      & 35.9     & 53.2       & 52.9      & 46.3           \\ 
    \midrule
    QLoRA~ ~                 & 4+16\&16                                                           & Yes                                                                       & Yes                                                                            & 46.0      & 37.3     & 55.8       & 55.1      & 48.4           \\
    IR-QLoRA~ ~              & 4+16                                                               & No                                                                        & No                                                                             & 47.2      & 39.0     & 56.5       & 55.0      & 49.3           \\
    IR-QLoRA(merged)        & 16                                                                 & No                                                                        & No                                                                             & 47.3      & 38.8     & 56.4       & 55.3      & 48.5           \\
    \textbf{Q-BLoRA (ours)}   & 4+16\&16                                                           & \textbf{Yes}                                                              & \textbf{Yes}                                                                   & 52.8      & 45.6     & 51.5       & 55.4      & \textbf{49.9}  \\ 
    \midrule
    PEQA                     & 4                                                                  & Yes                                                                       & Yes                                                                            & 43.0      & 37.7     & 53.6       & 49.0      & 45.0           \\
    QLoRA w/ GPTQ            & 4                                                                  & No                                                                        & Yes                                                                            & 45.4      & 37.4     & 55.7       & 54.3      & 48.0           \\
    QA-LoRA                  & 4                                                                  & No                                                                        & Yes                                                                            & 48.4      & 38.3     & 54.9       & 55.2      & 49.2           \\
    \textbf{QA-BLoRA (ours)}   & 4                                                                  & \textbf{Yes}                                                              & \textbf{Yes}                                                                   & 51.4      & 46.2     & 48.5       & 53.0      & \textbf{49.5}  \\ 
    \midrule
    LLaMA-33B                & 16                                                                 & -                                                                         & -                                                                              & 56.2      & 45.9     & 67.1       & 63.9      & 58.2           \\ 
    \midrule
    QLoRA~ ~ ~               & 4+16\&16                                                           & Yes                                                                       & Yes                                                                            & 55.4      & 46.0     & 66.4       & 63.6      & 57.7           \\
    IR-QLoRA~ ~              & 4+16                                                               & No                                                                        & No                                                                             & 56.7      & 46.7     & 66.5       & 63.2      & 58.2           \\
    \textbf{Q-BLoRA (ours)}   & 4+16\&16                                                           & \textbf{Yes}                                                              & \textbf{Yes}                                                                   & 61.4      & 52.2     & 60.7       & 64.2      & \textbf{59.0}  \\ 
    \midrule
    QA-LoRA                  & 4                                                                  & No                                                                        & Yes                                                                            & 55.8      & 46.4     & 67.0       & 64.0      & 58.1           \\
    \textbf{QA-BLoRA (ours)}   & 4                                                                  & \textbf{Yes}                                                              & \textbf{Yes}                                                                   & 61.6      & 52.4     & 57.7       & 63.2      & \textbf{58.3}  \\ 
    \midrule
    LLaMA-65B                & 16                                                                 & -                                                                         & -                                                                              & 61.4      & 51.9     & 73.6       & 67.6      & 63.4           \\ 
    \midrule
    QLoRA~ ~ ~               & 4+16\&16                                                           & Yes                                                                       & Yes                                                                            & 60.3      & 52.7     & 72.9       & 67.4      & 63.1           \\
    IR-QLoRA~ ~              & 4+16                                                               & No                                                                        & No                                                                             & 60.1      & 50.1     & 74.4       & 68.7      & 63.1           \\
    \textbf{Q-BLoRA (ours)}   & 4+16\&16                                                           & \textbf{Yes}                                                              & \textbf{Yes}                                                                   & 65.2      & 58.2     & 64.4       & 68.6      & \textbf{63.7}  \\ 
    \midrule
    QA-LoRA                  & 4                                                                  & No                                                                        & Yes                                                                            & 60.8      & 50.5     & 72.5       & 66.7      & 62.5           \\
    \textbf{QA-BLoRA (ours)}   & 4                                                                  & \textbf{Yes}                                                              & \textbf{Yes}                                                                   & 64.8      & 58.2     & 63.8       & 68.1      & \textbf{63.3}  \\
    \bottomrule
    \end{tabular}  
    }
    }
    \end{table*}

\textbf{Performance on LLaMA2.}
As shown in Table~\ref{table_llama2}, vanilla QLoRA with/without GPTQ exhibit remarkable model generalization capabilities. However, current SOTA variants (i.e., IR-QLoRA and QA-LoRA) show degraded performance. Only our Q-BLoRA and QA-BLoRA achieve accuracy superior to QLoRA. These results indicate the strong generalization across different LLM families for Q-BLoRA and QA-BLoRA.

    \begin{table}
    \centering
    \caption{Accuracy (\%) comparison of LLaMA2 on the MMLU fine-tuned on the Alpaca dataset. }
    \label{table_llama2}
    \resizebox{.99\columnwidth}{!}{
    \begin{tabular}{ccccccc} 
    \toprule
    \multirow{2}{*}{Method}          & \multirow{2}{*}{\#Bit} & \multicolumn{5}{c}{MMLU}                       \\ 
    \cline{3-7}
                                     &                        & Hums. & STEM & Social & Other & Avg.           \\ 
    \midrule
    LLaMA2-7B                        & 16                     & 43.0  & 36.4 & 51.4   & 52.2  & 45.5           \\ 
    \midrule
    QLoRA                            & 4+16\&16             & 49.1  & 36.4 & 52.6   & 54.1  & 47.0           \\
    IR-QLoRA                         & 4+16                   & 43.4  & 36.8 & 51.9   & 53.6  & 46.2           \\
    IR-QLoRA(merged)                 & 16                     & 42.4  & 36.5 & 51.2   & 52.9  & 45.6           \\
    \textbf{Q-BLoRA (ours)}           & 4+16\&16             & 50.0  & 42.5 & 51.3   & 51.2  & \textbf{48.2}  \\ 
    \midrule
    QLoRA w/ GPTQ                    & 4                      & 47.2  & 36.2 & 51.0   & 51.6  & 45.1           \\
    QA-LoRA                          & 4                      & 42.1  & 34.4 & 49.1   & 50.3  & 43.9           \\
    \textbf{QA-BLoRA (ours)}          & 4                      & 49.4  & 37.1 & 50.7   & 51.7  & \textbf{46.3}  \\ 
    \midrule
    LLaMA2-13B                       & 16                     & 53.3  & 44.1 & 63.3   & 61.0  & 55.3           \\ 
    \midrule
    QLoRA                            & 4+16\&16             & 55.8  & 44.0 & 63.6   & 61.0  & 55.6           \\
    IR-QLoRA                         & 4+16                   & 51.9  & 43.9 & 61.9   & 60.4  & 54.4           \\
    \textbf{\textbf{Q-BLoRA (ours)}}  & 4+16\&16             & 59.3  & 48.9 & 60.2   & 60.2  & \textbf{56.2}  \\ 
    \midrule
    QLoRA w/ GPTQ                    & 4                      & 53.7  & 43.6 & 62.6   & 60.0  & 53.8           \\
    \textbf{\textbf{QA-BLoRA (ours)}} & 4                      & 57.8  & 46.8 & 60.2   & 59.7  & \textbf{55.6}  \\
    \bottomrule
    \end{tabular}
    }
    \end{table}

\begin{figure}[htbp]
    	\centering
    	\subfloat[Results on Mistral-7B.]{\includegraphics[width=.485\columnwidth,height=.37\columnwidth]{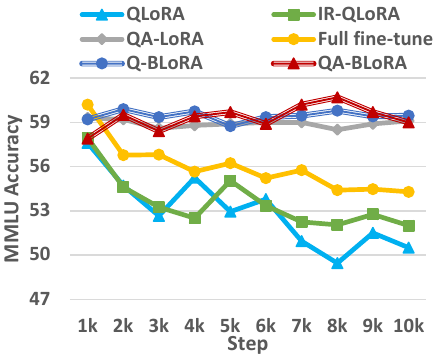}\label{fig_experiment_mistral}}\hspace{2pt}
    	\subfloat[Results on Gemma-2-2B.]{\includegraphics[width=.485\columnwidth,height=.37\columnwidth]{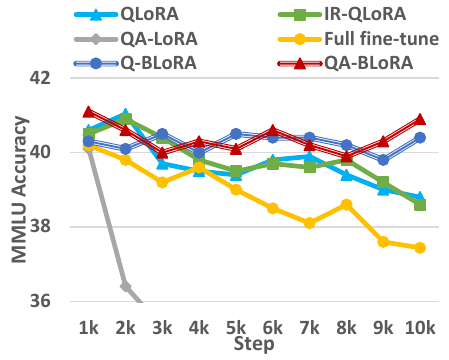}\label{fig_experiment_gemma}}\\
    	\caption{Comparison of the MMLU Accuracy (\%) during fine-tuning on the Alpaca dataset.}
            \label{fig_mistral_gemma}
    \end{figure}

\textbf{Performance on Mistral and Gemma.}
As shown in Figure~\ref{fig_mistral_gemma}, Q-BLoRA and QA-BLoRA demonstrate significant performance advantages on the latest Mistral-7B and Gemma-2-2B.

It is noteworthy that during fine-tuning, the performance of full fine-tune, QLoRA, and IR-QLoRA consistently degrades, while Q-BLoRA and QA-BLoRA maintain stable performance. This could be attributed to the fact that the latest models, such as Mistral and Gemma, have been pre-trained on high-quality data and diverse training strategies, with their model parameters carefully optimized \cite{jiang2023mistral, team2024gemma, huang2024good}. When further fine-tuned, significant parameter updates may disrupt the original model’s convergence. In contrast, Q-BLoRA and QA-BLoRA mitigate overfitting and extreme parameter updates by employing neutral operations on adapter inputs and outputs (i.e., \textsf{AvgPool} and \textsf{repeat\_interleave}), which helps preserve the model’s original performance.
While QA-LoRA applies similar compression operations and achieves good results on Mistral-7B, for the more parameter-efficient Gemma-2-2B, the lower dimensions of the adapter inputs lead to information loss and fine-tuning failure.

These results collectively demonstrate the effectiveness of Q-BLoRA and QA-BLoRA on the Mistral and Gemma models, aligning with the analysis presented in Section~\ref{sec_bara}.

\textbf{Fine-tune performance using Flan v2 dataset.}
We evaluated the fine-tuned models using the Flan v2 dataset to validate the performance of our methods on different datasets. As shown in Table~\ref{table_main_result_flanv2}, both our Q-BLoRA and QA-BLoRA achieve the highest accuracy in their respective application scenarios.

    \begin{table}
    \centering
    \caption{Accuracy (\%) comparison of LLaMA on the MMLU fine-tuned on the FLAN v2 dataset. }
    \label{table_main_result_flanv2}
    \resizebox{.99\columnwidth}{!}{
    \begin{tabular}{ccccccc} 
    \toprule
    \multirow{2}{*}{Method } & \multirow{2}{*}{\#Bit} & \multicolumn{5}{c}{MMLU    }          \\ 
    \cline{3-7}
                             &                        & Hums. & STEM & Social & Other & Avg.  \\ 
    \midrule
    LLaMA-7B                 & 16                     & 33.3  & 29.8 & 37.8   & 38.0  & 34.6  \\
    \midrule
    QLoRA                    & 4+16\&16               & 41.4  & 35.0 & 49.8   & 52.0  & 44.3  \\
    IR-QLoRA                 & 4+16                   & 44.2  & 39.3 & 54.5   & 52.9  & 47.4  \\
    Q-BLoRA (ours)            & 4+16\&16               & 50.8  & 40.2 & 48.2   & 52.8  & 47.6  \\ 
    \midrule
    QLoRA w/ GPTQ            & 4                      & 36.5  & 33.7 & 46.9   & 50.3  & 41.4  \\
    QA-LoRA                  & 4                      & 43.9  & 38.0 & 54.3   & 53.0  & 47.0  \\
    QA-BLoRA (ours)           & 4                      & 50.6  & 40.0 & 48.1   & 52.3  & 47.2  \\
    \bottomrule
    \end{tabular}
    }
    \end{table}

\textbf{Performance on Commonsense QA.}
We also evaluate our Q-BLoRA and QA-BLoRA for zero-shot commonsense QA, with results summarized in Table~\ref{table_commonqa}. The results are very similar to those on MMLU. Our Q-BLoRA consistently achieves the SOTA performance, and our QA-BLoRA also significantly outperforms other methods on 4-bit inference models. Notably, on LLaMA2, only our methods surpass QLoRA and the 4-bit model obtained by QLoRA w/GPTQ, while other methods fail to exceed the benchmarks.

\begin{table*}
\centering
\caption{Accuracy (\%) comparison on the 0-shot commonsense QA. }
\label{table_commonqa}
\setlength{\tabcolsep}{4mm}{
\resizebox{.97\textwidth}{!}{
\begin{tabular}{cccccccccc} 
\toprule
Method                           & \#Bit    & HellaSwag & ~ARC-e~ & ~ARC-c~ & ~PIQA~ & WinoGrande & ~BoolQ~ & ~OBQA~ & ~Avg.~         \\ 
\midrule
LLaMA-7B                         & 16       & 56.3      & 67.3    & 38.2    & 78.2   & 67.1       & 72.9    & 28.4   & 58.3           \\ 
\midrule
QLoRA                            & 4+16\&16 & 61.8      & 75.8    & 43.6    & 78.1   & 68.4       & 73.7    & 32.8   & 62.0           \\
IR-QLoRA                         & 4+16     & 54.7      & 76.6    & 45.1    & 78.8   & 72.6       & 80.6    & 37.2   & 63.7           \\
\textbf{Q-BLoRA (ours)}           & 4+16\&16 & 76.3      & 57.2    & 43.8    & 78.7   & 70.2       & 78.4    & 60.4   & \textbf{66.4}  \\ 
\midrule
QLoRA w/ GPTQ                    & 4        & 57.4      & 70.9    & 41.8    & 77.6   & 66.2       & 73.5    & 31.2   & 59.8           \\
QA-LoRA                          & 4        & 58.6      & 71.2    & 43.9    & 78.0   & 66.9       & 79.9    & 43.0   & 61.8           \\
\textbf{QA-BLoRA (ours)}          & 4        & 76.3      & 56.6    & 43.4    & 79.2   & 70.2       & 77.9    & 58.2   & \textbf{66.0}  \\ 
\midrule
LLaMA2-7B                        & 16       & 75.0      & 60.4    & 40.3    & 75.2   & 65.8       & 75.2    & 65.6   & 65.4           \\ 
\midrule
QLoRA                            & 4+16\&16 & 75.6      & 68.9    & 45.7    & 77.8   & 68.6       & 77.2    & 72.2   & 69.4           \\
IR-QLoRA                         & 4+16     & 75.4      & 68.9    & 45.5    & 77.0   & 68.4       & 77.0    & 69.4   & 68.8           \\
\textbf{\textbf{Q-BLoRA (ours)}}  & 4+16\&16 & 76.3      & 70.0    & 46.8    & 78.6   & 70.2       & 77.5    & 69.4   & \textbf{69.8}  \\ 
\midrule
QLoRA w/ GPTQ                    & 4        & 75.1      & 66.2    & 44.5    & 76.2   & 67.1       & 73.5    & 69.4   & 67.4           \\
QA-LoRA                          & 4        & 75.0      & 65.6    & 44.2    & 75.8   & 66.7       & 73.2    & 69.2   & 67.1           \\
\textbf{\textbf{QA-BLoRA (ours)}} & 4        & 75.6      & 65.8    & 48.8    & 78.7   & 69.9       & 77.5    & 65.4   & \textbf{68.8}  \\
\bottomrule
\end{tabular}
}
}
\end{table*}

\textbf{Evaluations using GPT-4/ChatGPT.}
We present the evaluation results of content generated by Q-BLoRA and QA-BLoRA compared to QLoRA, using GPT-4/ChatGPT as the evaluator. This evaluation simulates human judgment, offering unique insights that complement automated metrics and provide an alternative perspective on the performance of fine-tuned models.
Each response is scored by GPT-4/ChatGPT on a 1–10 scale. To mitigate positional bias, the responses are presented in two different orders. The evaluation criteria are defined as follows:
\begin{itemize} [leftmargin=*, topsep=0pt]
\setlength{\topsep}{0pt} 
\setlength{\itemsep}{0pt} 
\setlength{\parsep}{0pt} 
\setlength{\parskip}{0pt} 
\item Win: Outperforms in both orderings or wins in one and ties in the other.
\item Tie: Ties in both orderings or wins in one and loses in the other. 
\item Loss: Performs worse in both orderings or ties in one and loses in the other. 
\end{itemize}

\begin{figure}[htbp]
    	\centering
    	\subfloat[Q-BLoRA vs. QLoRA.]{\includegraphics[width=.48\columnwidth,height=.3\columnwidth]{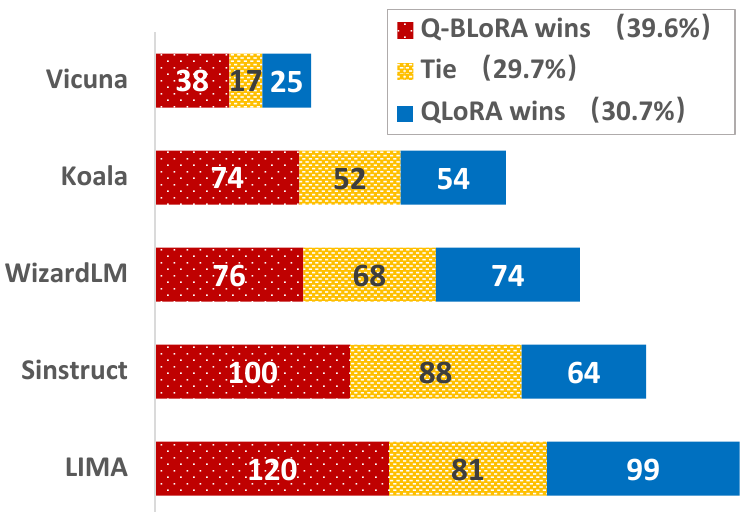}\label{fig_gpt_qblora}}\hspace{5pt}
    	\subfloat[QA-BLoRA vs. QLoRA.]{\includegraphics[width=.48\columnwidth,height=.3\columnwidth]{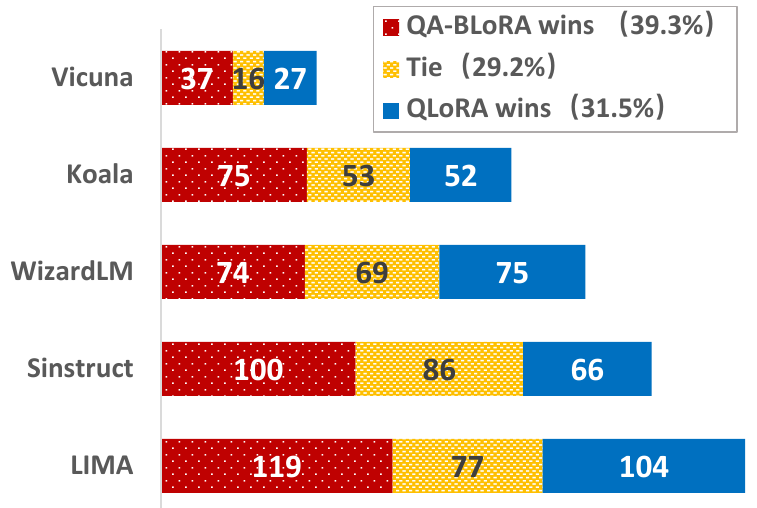}\label{fig_gpt_qablora}}\\
    	\caption{Comparison of GPT-4/ChatGPT evaluations.}
            \label{fig_gpt}
    \end{figure}

The Alpaca dataset and LLaMA-7B are used for fine-tuning in this evaluation. The results in Figure~\ref{fig_gpt} demonstrate that our proposed Q-BLoRA and QA-BLoRA outperform QLoRA in nearly all tests, highlighting the superior performance of Q-BLoRA and QA-BLoRA.

\textbf{The efficiency of Q-BLoRA and QA-BLoRA.}
Table~\ref{table_efficiency} compares the memory usage, fine-tuning time, and accuracy of various fine-tuning methods. The key findings are as follows: (i) Compared to non-quantized fine-tuning, both Q-BLoRA and QA-BLoRA significantly reduce the memory required for fine-tuning while maintaining similar accuracy. (ii) Compared to Q-LoRA, Q-BLoRA not only improves accuracy under similar fine-tuning times but also outperforms IR-QLoRA in terms of both accuracy and fine-tuning time, as well as inference efficiency. (iii) QA-BLoRA further reduces the size of the inference model. Compared to QLoRA with GPTQ, QA-BLoRA achieves a significant improvement in accuracy without requiring additional PTQ operations, and the fine-tuning time is shorter. When compared to QA-LoRA, QA-BLoRA shows significant advantages in both accuracy and fine-tuning time. (iv) Additionally, the results show that non-quantized BLoRA does not lead to significant accuracy improvements compared to full fine-tuning or LoRA. However, in the case of quantization, our methods significantly improve memory efficiency while maintaining comparable performance. These results indicate that, in the non-quantized case, no underfitting occurs, while in the quantized case, our methods alleviate underfitting and thus achieve better model performance.

These results demonstrate the superior efficiency of Q-BLoRA and QA-BLoRA.


    \begin{table}
    \centering
    \caption{Efficiency comparison: block size (BS), memory requirements for fine-tuning (Mem-FT) and inference (Mem-Infer.)(e.g., "13.5+4.9G" indicates 13.5GB for the pre-trained model and 4.9GB for the adapter), fine-tuning time (Time), and accuracy on MMLU with Alpaca (Acc.). All results are reported with one GeForce RTX 3090 GPU.}
    \label{table_efficiency}
    \resizebox{.99\columnwidth}{!}{
    \begin{tabular}{cccccc} 
\toprule
Method             & BS & Mem-FT    & Mem-Infer. & Time (h) & ~Acc.~         \\ 
\midrule
Full fine-tune     & -           & 13.5+32G  & 13.5G      & -        & 41.0           \\
LoRA               & -           & 13.5+4.9G & 13.5G      & 10.7     & 41.3           \\
BLoRA              & -           & 13.5+4.9G & 13.5G      & 10.8     & 41.3           \\ 
\midrule
QLoRA              & 32          & 3.8+4.9G  & 13.5G      & 13.1     & 38.4           \\
IR-QLoRA           & 32          & 3.8+4.9G  & 13.5+0.2G  & 14.0     & 40.8           \\
\textbf{Q-BLoRA~}  & 32          & 3.8+4.9G  & 13.5G      & 13.2     & \textbf{41.4}  \\ 
\midrule
QLoRA w/GPTQ      & 32          & 3.8+4.9G  &  3.8G      & 13.1+0.1 & 36.0           \\
QA-LoRA            & 32          & 3.8+2.7G  & 3.8G       & 0.1+15.8     & 36.4           \\
\textbf{QA-BLoRA~} & 32          & 3.8+4.9G  & 3.8G       & 13.0     & \textbf{40.7}  \\
\textbf{QA-BLoRA}  & 64          & 3.8+2.5G  & 3.8G       & 12.8     & \textbf{38.2}  \\
\bottomrule
\end{tabular}
    }
    \end{table}

\subsection{Ablation Study}
\label{sec_ablation}

\begin{figure}[htbp]
    	\centering
    	\subfloat[Impact of the rank and the scale factor of Q-BLoRA.]{\includegraphics[width=.48\columnwidth,height=.38\columnwidth]{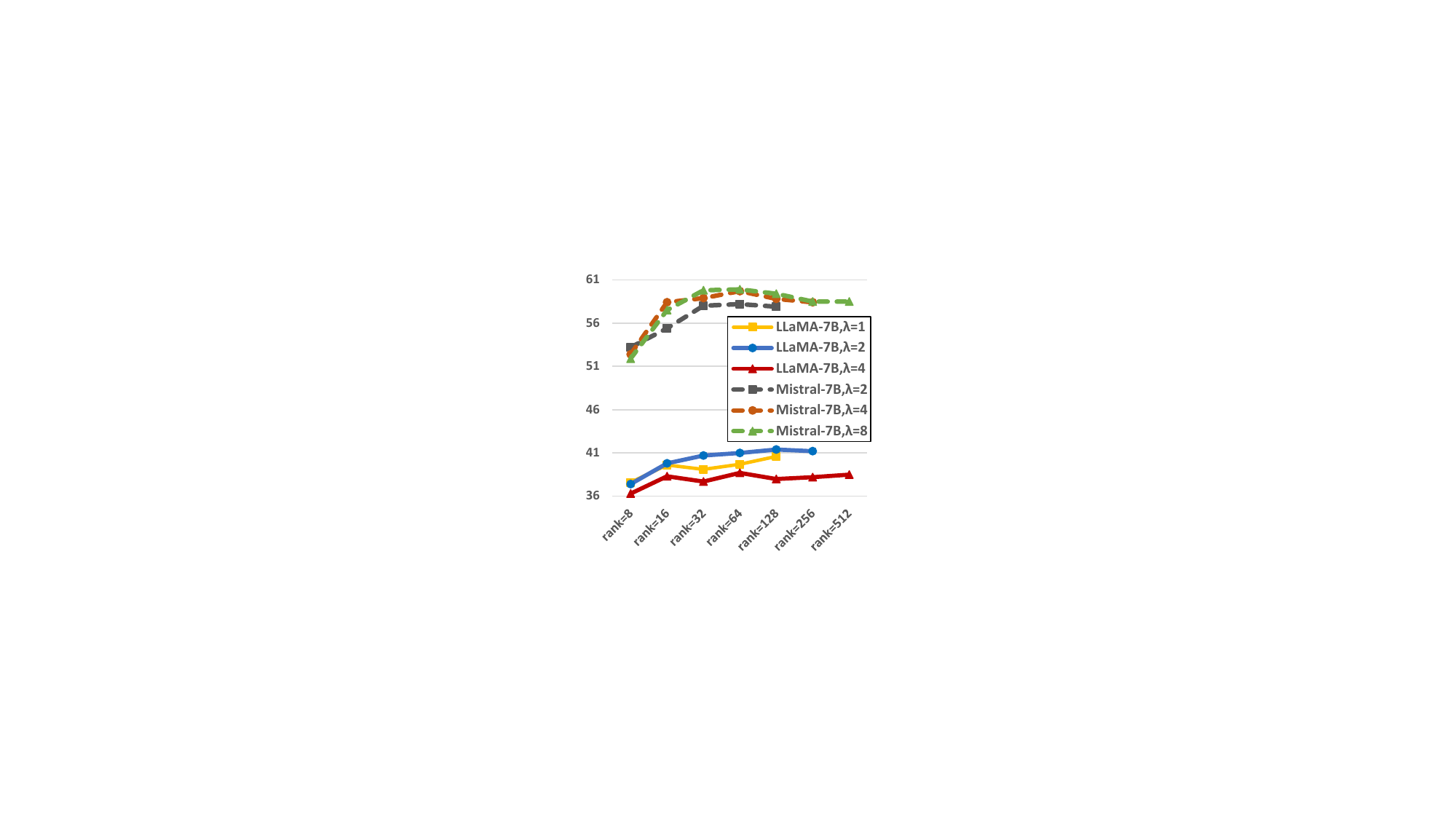}\label{fig_ablation1}}\hspace{5pt}
    	\subfloat[Impact of the scale factor of QA-BLoRA.]{\includegraphics[width=.48\columnwidth,height=.38\columnwidth]{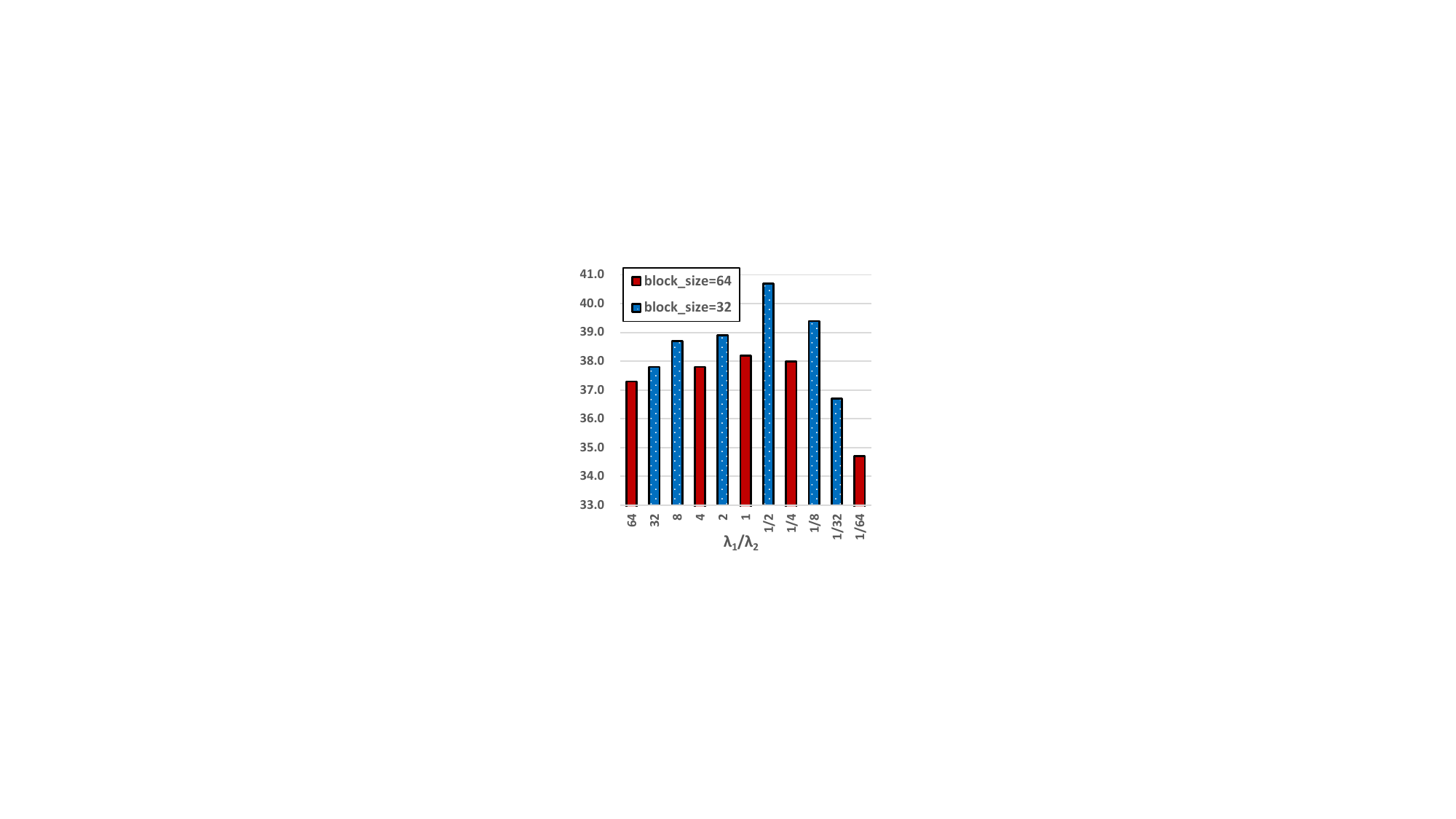}\label{fig_ablation2}}\\
    	\caption{Ablation analysis.}
            \label{fig_ablation}
    \end{figure}

\textbf{The balancing factor of Q-BLoRA}. Q-BLoRA introduces a new model-specific hyperparameter $\lambda$. As shown in Figure~\ref{fig_ablation1}, when fine-tuning LLaMA-7B, setting 
$\lambda=2$ yields better performance, while for Mistral-7B, $\lambda=8$ achieves superior results. We determine the value through experiments and recommend $\lambda=2$ for the LLaMA/LLaMA2 model family and $\lambda=8$ for Mistral and Gemma models.

As discussed in Figure~\ref{fig_mistral_gemma}, the SOTA LLMs benefit from high-quality data and meticulously designed training strategies during pre-training, leading to well-converged model parameters. Consequently, conventional fine-tuning approaches may compromise the original model's performance. Referring to the analysis in Figure~\ref{fig_underfit}, our method effectively suppresses drastic parameter updates. By employing a larger $\lambda$ for SOTA LLMs, we can further preserve the original performance, thereby mitigating the negative impact of regular fine-tuning. Additionally, we observe that using a lower learning rate also helps restrain parameter updates and alleviates performance degradation to some extent. Exploring how to integrate our method to develop optimized fine-tuning strategies for SOTA LLMs will be a focus of future work.




\textbf{The balancing factor of QA-BLoRA.}
In QA-BLoRA, the values of $\lambda_1$ and $\lambda_2$ are aligned with the block quantization of the pre-trained model. In LLMs, the block size for quantization is typically set to 32 or 64 \cite{dettmers2023case,xu2023qa}. Figure~\ref{fig_ablation2} shows the fine-tuning results for different quantization blocks (with corresponding values of $\lambda_1$ and $\lambda_2$) under block sizes of 32 and 64. It can be observed that when the block size is 64, the best performance is achieved with $\lambda_1 = \lambda_2 = 8$, whereas for a block size of 32, the best performance occurs with $\lambda_1 = 4$ and $\lambda_2 = 8$.



\textbf{The scale operators.}
We use alternative scale operators to study the effects of other non-parameter operators. Besides the \textsf{AvgPool} \& \textsf{repeat\_interleave}, we also tried \textsf{truncation} \& \textsf{supplement\_0} and \textsf{interval\_sampling} \& \textsf{interpolation}. The specific descriptions are as follows:

    \begin{itemize}[leftmargin=*, topsep=0pt]
    \setlength{\itemsep}{0pt}
    \setlength{\leftmargin}{0pt}
        \item \textsf{AvgPool} \& \textsf{repeat\_interleave}: see Section~\ref{sec_bara}.
        \item \textsf{truncation} \& \textsf{supplement\_0}: Truncate the input to retain only the front part; restore the output to the original size using zero-padding.
        \item \textsf{interval\_sampling} \& \textsf{interpolation}: Sample values at regular intervals; restore the output to the original size using zero-interpolation.
    \end{itemize}

As shown in Table~\ref{table_ablation_operator}, other scaling operators are significantly less effective compared to our \textsf{AvgPool} \& \textsf{repeat\_interleave} operator under different settings. This is due to the presence of outliers in certain channels, which carry high information content. Completely discarding a portion of outliers has a significant impact on performance. Furthermore, as discussed in Section~\ref{sec_bara}, the neutral operation of \textsf{AvgPool} \& \textsf{repeat\_interleave} reduces the gap between values across different channels, thereby lowering the difficulty of fitting. In contrast, other operators not only lose valuable information from outliers, but also fail to effectively balance the values across channels.

\begin{table}
\centering
\caption{Accuracy (\%) comparison for operators on the MMLU fine-tuned on the Alpaca dataset. }
\label{table_ablation_operator}
\resizebox{.99\columnwidth}{!}{
\begin{tabular}{ccccc} 
\toprule
\multirow{2}{*}{Operator}           & \multicolumn{2}{c}{Q-BLoRA}                          & \multicolumn{2}{c}{QA-BLoRA}                                                                                       \\ 
\cline{2-5}
                                    & $\lambda$= 2 & $\lambda$=4 & $\lambda_1$=4,$\lambda_2$=8 & $\lambda_1$=$\lambda_2$=8  \\ 
\midrule
AvgPool\&repeat\_interleave       & \textbf{41.4}                     & \textbf{38.2}                     & \textbf{40.7}                                                     & \textbf{38.2}                                                   \\
truncation\&supplement\_0           & 36.6                     & 35.8                     & 37.0                                                     & 35.9                                                   \\
int\_sampling\&interpolation & 36.8                     & 36.4                     & 36.8                                                     & 36.0                                                   \\
\bottomrule
\end{tabular}
}
\end{table}


\section{Conclusion}
In this paper, we propose two methods for fine-tuning quantized LLMs: Q-BLoRA and QA-BLoRA. Q-BLoRA addresses underfitting during fine-tuning by balancing the complexity of the adapter's input and output with the rank of the adapter. Additionally, it enables lossless merging of adapter parameters after fine-tuning, resulting in high-precision floating-point inference models. Building on the parameter merging approach of Q-BLoRA, QA-BLoRA achieves quantization-aware fine-tuning of low-rank adaptation by aligning with the block-wise quantization of the pre-trained model. This allows for the direct acquisition of low-precision inference models. Both Q-BLoRA and QA-BLoRA are easy to implement and demonstrate significant superiority across various models and evaluation tasks.

\section*{Acknowledgments}
We would like to thank the anonymous reviewers and the action editor for their insightful comments.
This work is sponsored in part by the National Natural Science Foundation of China (No. 62025208 and 62421002).

\end{document}